\pdfoutput=1
\documentclass[10pt,twocolumn,letterpaper]{article}

\usepackage{iccv}
\usepackage{times}
\usepackage{epsfig}
\usepackage{graphicx}
\usepackage{amsmath}
\usepackage{amssymb}
\usepackage{booktabs}
\usepackage{enumitem}
\usepackage{caption}
\usepackage{subcaption}
\usepackage{authblk}

\usepackage[pagebackref=true,breaklinks=true,letterpaper=true,colorlinks,bookmarks=false]{hyperref}

\usepackage[capitalize]{cleveref}
\crefname{section}{Sec.}{Secs.}
\Crefname{section}{Section}{Sections}
\Crefname{table}{Table}{Tables}
\crefname{table}{Tab.}{Tabs.}

\newcommand{\Dn}[1]{\textcolor{red}{$_{\downarrow#1}$}}
\newcommand{\Up}[1]{\textcolor{blue}{$_{\uparrow#1}$}}
\iccvfinalcopy 

\def\iccvPaperID{9019} 

\ificcvfinal\fi

\begin{document}

\title{De-coupling and De-positioning Dense Self-supervised Learning}

\setlength{\affilsep}{0.2em}
\author[1]{Congpei Qiu\thanks{The first two authors contributed equally to this work.}}
\author[2]{Tong Zhang$^{\scriptsize*}$}
\author[1]{Wei Ke}
\author[2]{Mathieu Salzmann}%
\author[2]{Sabine Süsstrunk}
\affil[1]{School of Software Engineering, Xi'an Jiaotong University, China}
\affil[2]{School of Computer and Communication Sciences, EPFL, Switzerland }
\renewcommand*{\Authands}{, }
\maketitle
\ificcvfinal\thispagestyle{empty}\fi
\begin{abstract}


Dense Self-Supervised Learning (SSL) methods address the limitations of using image-level feature representations when handling images with multiple objects. Although the dense features extracted by employing segmentation maps and bounding boxes allow networks to perform SSL for each object, we show that they suffer from coupling and positional bias, which arise from the receptive field increasing with layer depth and zero-padding. We address this by introducing three data augmentation strategies, and leveraging them in (i) a decoupling module that aims to robustify the network to variations in the object's surroundings, and (ii) a de-positioning module that encourages the network to discard positional object information. We demonstrate the benefits of our method on COCO and on a new challenging benchmark, OpenImage-MINI, for object classification, semantic segmentation, and object detection. Our extensive experiments evidence the better generalization of our method compared to the SOTA dense SSL methods.\footnote{Our code, models, and data will be made publicly available at \href{https://github.com/ztt1024/denseSSL} {https://github.com/ztt1024/denseSSL}. Correspondence to Wei Ke.}
\end{abstract}


\section{Introduction}
\label{sec:intro}
Self-supervised learning (SSL) has greatly progressed, reaching the point where it significantly outperforms traditional supervised pre-training when transferring to different domains and tasks in the presence of small amounts of supervised data~\cite{chaitanya2020contrastive, ericsson2021well}. SSL seeks to maximize the feature similarity between positive pairs~\cite{chen2020simple,chen2021exploring, grill2020bootstrap}, which are constructed by applying two random distinctive augmentations to the same image. Unfortunately, in the presence of challenging data, such as images containing multiple objects with different sizes, or imbalanced numbers of images across the semantic classes, such a pairing mechanism will generate positive pairs containing different semantic objects, thus confusing the learning process.

\begin{figure}[t]
  \centering
  \includegraphics[width=1.0\linewidth]{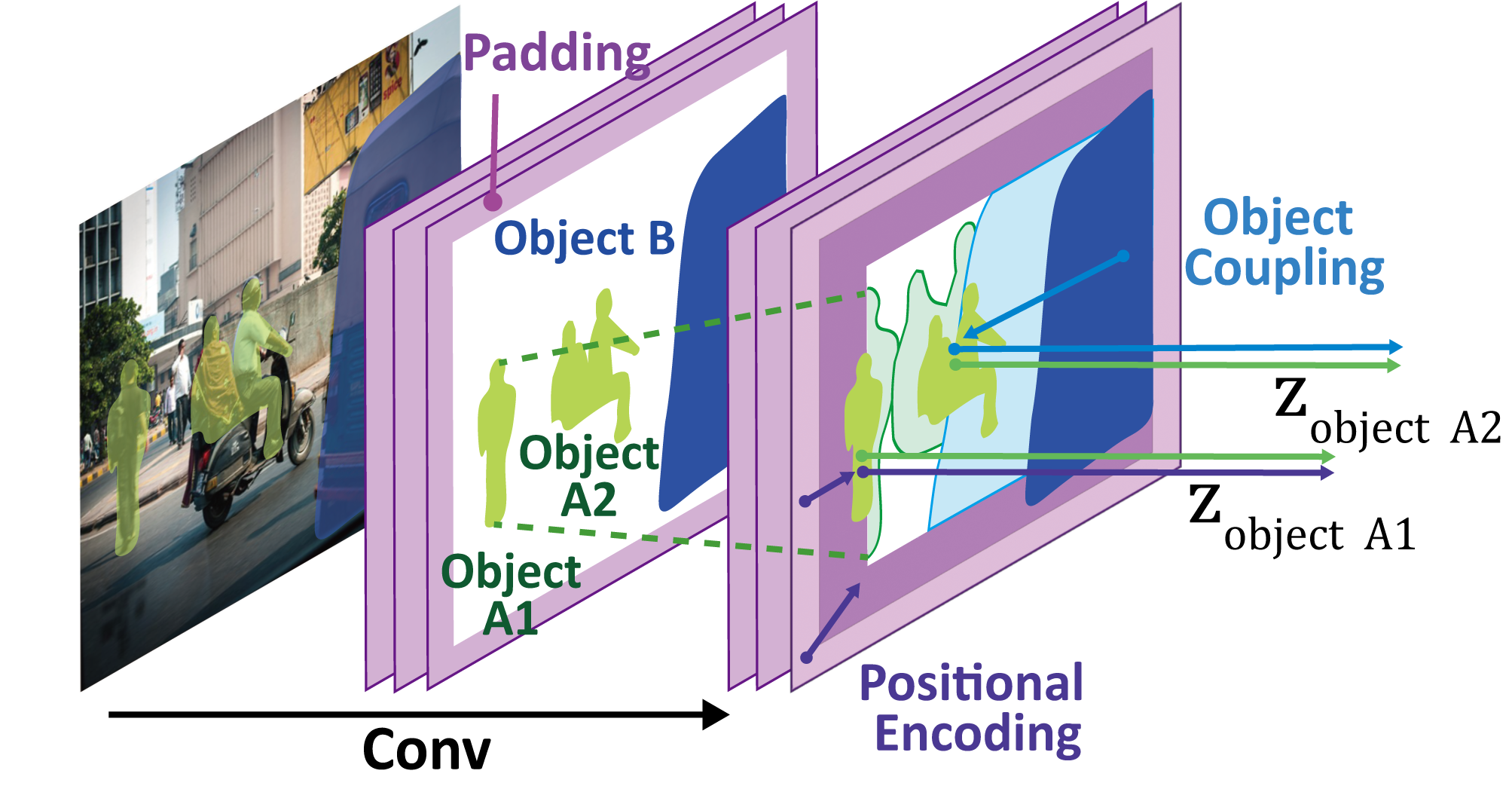}
  \vspace{-0.6cm}
  \caption{{\bf Object coupling and positional bias} on the region-level features extracted from a scene with multiple objects. The overlap between objects increases with the depth within the network, thus mixing information across objects. Zero-padding inherently encodes a positional bias in the features, thus leading to different features for the same object seen at two different locations. 
  }
  \vspace{-0.6cm}
  \label{fig:problem_overview}
\end{figure}


This problem has nonetheless been studied in the recent literature. In particular, existing dense SSL methods aim to find either region-level~\cite{henaff2021efficient, wei2021aligning}, or point-level~\cite{bai2022point,wang2021dense,wang2022exploring, xie2021propagate}  correspondences to construct positive and negative pairs. These dense SSL methods are based on the intuition that features from different objects should be learned separately in order to preserve all the object features, rather than focusing only on discriminative objects.
In this paper, however, we show that SSL for such positive pairs is inherently perturbed by factors related to the \textit{structure} of the neural network itself and by the  use of \textit{homogeneous augmentations} applied to all objects in one view. We highlight these two factors in Fig.~\ref{fig:problem_overview}: (i) In the deeper, lower-resolution network layers, features of objects become coupled, entangling information from neighboring objects; (ii) The use of zero-padding implicitly encodes a positional bias in the features, making the features extracted from the exact same image patch at different spatial locations differ.

\begin{figure}[t]
  \centering
  \includegraphics[width=1.0\linewidth]{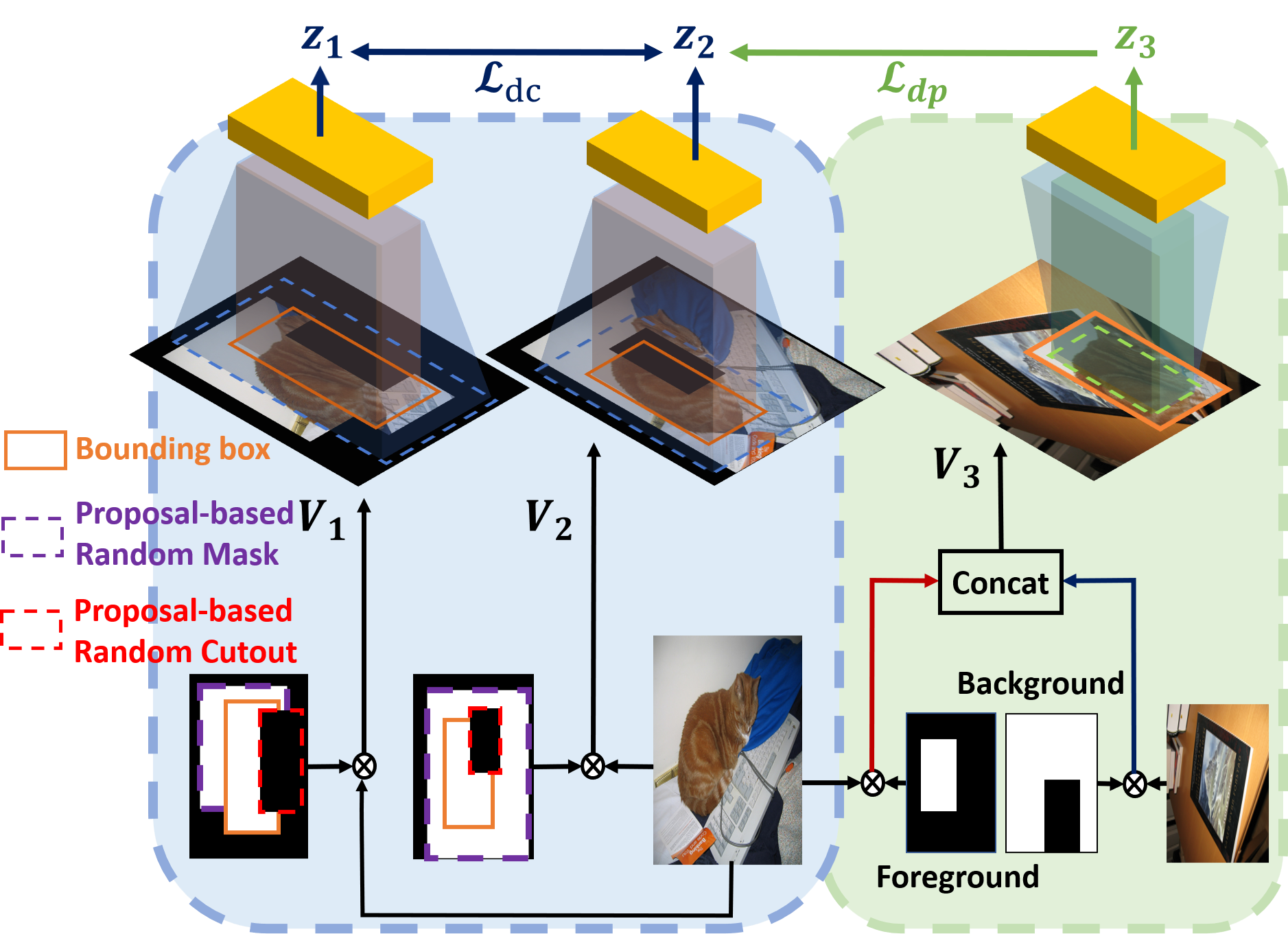}
  \vspace{-0.6cm}
  \caption{{\bf Overview of our method.} We introduce novel region-level augmentations and apply them to the bounding boxes in every view to construct positive pairs. The de-coupling module employs positive pairs that have distinct cut-out regions and backgrounds of the same bounding box to encourage the network to be robust to 
  the coupling of nearby objects. The de-positioning module builds positive pairs by placing an object region at different spatial locations with a different background, thus forcing the network to discard the positional bias and the surroundings.}

  
 \vspace{-0.6cm}
  \label{fig:method_overview}
\end{figure}

In this paper, we introduce a simple yet effective method, namely \textbf{D}e-coupling and \textbf{D}e-positioning \textbf{D}ense \textbf{SSL} (D$^3$SSL), to address the above issues. D$^3$SSL introduces three novel region-level augmentations, which generate distinct views for every object in the same image, and exploits these augmentations in two network modules, as shown in Fig.~\ref{fig:method_overview}. 
Specifically, the \textit{de-coupling module} leverages positive region pairs that are composed by applying two independent box jitterings to the same bounding box, considering a different amount of background, and cutting out different portions of the bounding box for each view. These boxes are then fed to a teacher encoder, whose goal is to learn view-invariant representations. In contrast, the \textit{de-positioning} module randomly positions a bounding box at two different locations, blending it with completely different backgrounds. This yields views with higher variance than one in the de-coupling module, which are then fed to a student encoder. The goal is to enforce the same representation as the de-coupling module while avoiding confusing it. Note that the de-positioning module also contributes to de-coupling. That is, these two modules aim to decouple the learned features from the object's surroundings and their spatial locations, enabling self-supervised training on complex scenes with multiple objects. 

Our contributions can be summarized as follows:
\begin{enumerate}[topsep=0.2pt, itemsep=0.5pt, parsep=2pt]
    \item We empirically demonstrate the negative impact of the coupling of objects and the positional bias on dense SSL methods.
    
    \item We develop three novel region-level augmentations and introduce two modules that exploit these augmentations to overcome these challenges. We further validate our novel augmentations on SoCo~\cite{wei2021aligning}.
    
    \item  We propose object-level k-nearest neighbors (KNN) as an evaluation technique for dense SSL methods. It directly reflects the quality of the object features without requiring any fine-tuning.
\end{enumerate}

Moreover, we also create a challenging dataset, OpenImage-MINI, a subset of OpenImage~\cite{OpenImages2, OpenImages}. It contains on average 9.1 objects per image and has imbalanced numbers of objects for each category, which opens the door to developing and benchmarking dense SSL methods in the wild. Our code and dataset will be made publicly available upon acceptance of the paper.

\section{Related Work}\label{sec:related}

\textbf{Image-level Self-supervised Learning.} In the context of deep learning, SSL enables pre-training without supervision~\cite{dosovitskiy2014discriminative}. The first attempts at doing so involved learning auxiliary tasks such as colorization~\cite{zhang2016colorful}, solving jigsaw puzzles~\cite{noroozi2016unsupervised}, and predicting  rotation~\cite{gidaris2018unsupervised}. Although these pretext tasks allow the network to learn meaningful representations, their effectiveness is sensitive to the network architecture~\cite{kolesnikov2019revisiting} and the downstream task.

With the introduction of InfoNCE~\cite{oord2018representation}, inspired by the triplet loss in metric learning~\cite{oh2016deep,sohn2016improved}, contrastive learning has become the dominating strategy in SSL. In this context, SimCLR~\cite{chen2020simple} constructs positive pairs by applying random data augmentation to the same image twice, while treating the other images as negative pairs; MoCo~\cite{he2020momentum} incorporates an additional memory bank to obtain more negative samples, which stabilizes the training process; BYOL~\cite{grill2020bootstrap} argues, by contrast, that negative pairs are not necessary, thanks to the help of asymmetric network updates; and SimSiam~\cite{chen2021exploring} further simplifies the network by replacing the momentum encoder with a stop-gradient strategy. Other approaches to contrastive SSL include clustering-based methods~\cite{asano2019self, caron2018deep,caron2020unsupervised,clustering2,zhang2019neural,clustering4} and feature decorrelation ones~\cite{zbontar2021barlow}. 
In any event, these methods focus on capturing semantic invariance via \emph{image-level} augmentation. This approach, however, is ill-suited to handle images with multiple objects~\cite{he2019rethinking,zhang2022leverage}, for which locality must be considered.


\textbf{Dense Self-supervised Learning.} Dense SSL was proposed to pre-train on complex datasets, such as COCO~\cite{lin2014microsoft}. Existing methods typically use the same loss functions as image-level ones, but introduce new ways of building positive pairs on finer scales. In particular, the point-level methods, such as DenseCL~\cite{wang2021dense}, PixPro~\cite{xie2021propagate}, and VADeR~\cite{o2020unsupervised}, construct positive pairs by calculating the pairwise cosine distances of all the points from two augmented views and selecting the ones having higher similarity. SetSim~\cite{wang2022exploring} improves the selecting stage by using set similarity to discard the noisy pairs. Point-level RCL~\cite{bai2022point} manually divides an image into grids to construct the positive point pairs within the same grid and performs affinity distillation for all the points. However, all of the above methods rely on image-level SSL as initialization to find matching points. As a result, they inherit the shortcomings of image-level SSL, which leads to degraded performance on datasets with multiple objects per image. Moreover, the point-level methods are expensive in terms of both memory and computation as they compute the loss for all the point pairs.

To mitigate this issue, Self-EMD~\cite{liu2020self} employs region-level information to pre-train the network using the Earth mover's distance on pairs of feature maps. Concurrently, InsLoc~\cite{yang2021instance} utilizes pretext tasks to facilitate region-level contrastive learning, DetCon~\cite{henaff2021efficient} employs unsupervised masks,  and SoCo~\cite{wei2021aligning} uses selective search~\cite{uijlings2013selective} bounding boxes as external labor-free supervision to build the positive pairs. 
However, all the existing dense methods ignore that objects' features are affected by their neighbors and spatial positions in the image.
Here, we show that this has the undesirable effects of coupling adjacent objects and creating a positional bias, and introduce novel region-level augmentations and network modules to address these issues.



\section{Motivation: Empirical Analysis}\label{sec:ana}
We motivate with an empirical analysis highlighting the challenges that need to be addressed in dense SSL. As shown in Fig.~\ref{fig:problem_overview}, due to the nature of convolution neural networks, the dense-level features are largely affected by the coupling of objects and a positional bias. In this section, we present the outcomes of our empirical analysis, quantifying these two issues. We defer  the experimental details and additional experiments 
about the behavior of image-level loss on images with multiple objects to the supplementary material. In the following experiments, the baseline point-level method simply uses absolute coordinates to create the positive pairs; for the other evaluated methods, we used the official checkpoints provided by their respective authors.





\vspace{-0.1cm}
\subsection{Objects Coupling}
\label{sec:coupling}

\begin{figure}[t]
  \centering
  \begin{subfigure}[b]{0.47\linewidth}
    \centering
        \includegraphics[width=1\linewidth]{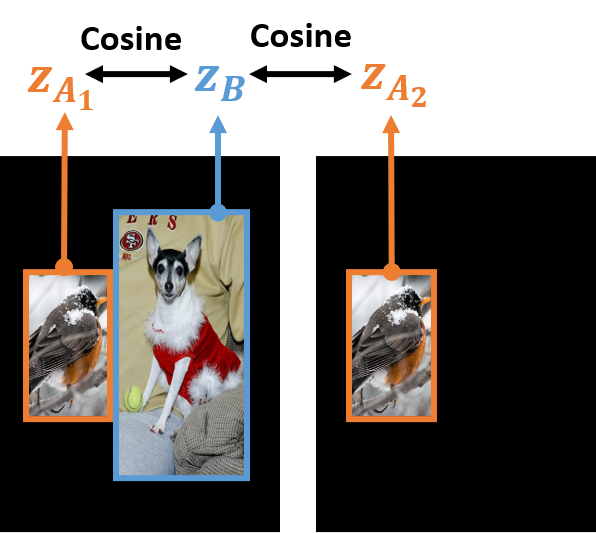}
    \caption{}
    \label{fig:coupling-a}
    \vspace{-0.3cm}
  \end{subfigure}
  \hfill
  \begin{subfigure}[b]{0.5\linewidth}
  \centering
        \includegraphics[width=1\linewidth]{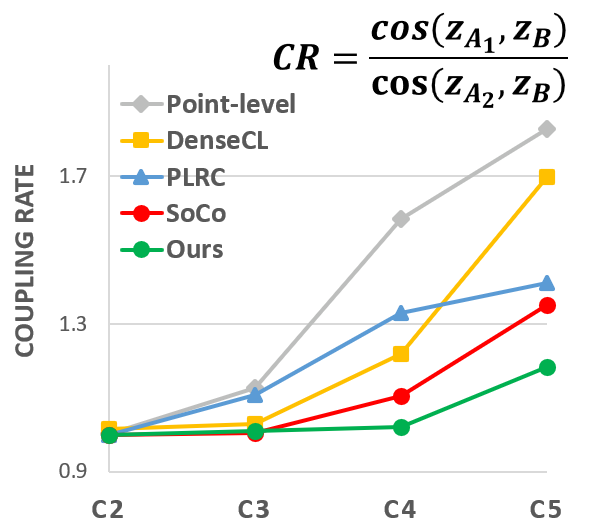}
    \caption{}
    \label{fig:coupling-b}
    \vspace{-0.3cm}
  \end{subfigure}
  \caption{ \textbf{Measuring object coupling.} (a) We create one image containing object A and object B, and the other containing object A at the same position as in the first image. (b) We plot the coupling rate (CR) for different layers of the network. Note that deeper layers are more strongly affected by object coupling because of their larger receptive fields.}
 \vspace{-0.5cm}
  \label{fig:coupling}
\end{figure}

Let us now study the problem of object coupling, due to the feature correlation between two neighboring objects in an image. To this end, we consider pairs of objects (A and B) belonging to different categories. Then, as illustrated in Fig.~\ref{fig:coupling-a}, we synthesize one image that contains both objects next to each other in front of a black background, and one image that only contains object A at the same position as in the other image. We then compute the cosine similarity between the features extracted from A1 (object A in the first image) and from B, and the one between the features extracted from A2 (object A in the second image) and from B. We denote the ratio of these two similarities as the coupling rate (CR). In an ideal scenario, the coupling rate should be 1, indicating that the presence of B does not affect the features extracted from A. In Fig.~\ref{fig:coupling-b}, we plot the coupling rate for different layers of the network. As the features are extracted in the deeper network layer, the coupling rate obtained by different SSL methods increases, confirming that the larger receptive field of these layers mixes information from both objects. Point-level methods (point-level, DenseCL, and PLRC) suffer from more severe coupling than region-level ones (SoCo and Ours).  



\subsection{Positional Bias}
\label{sec:positioning}
\begin{figure}[h]
  \centering
  \begin{subfigure}[b]{0.43\linewidth}
    \centering
        \includegraphics[width=1\linewidth]{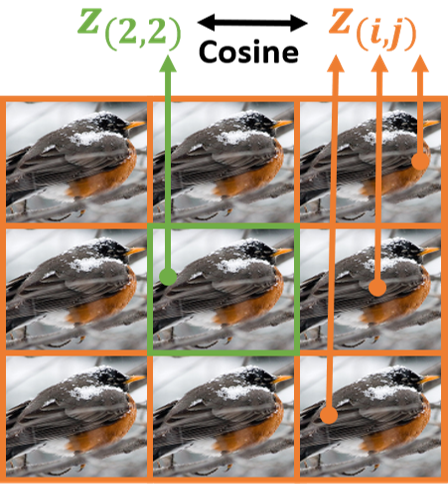}
    \caption{}
    \label{fig:position-a}
  \end{subfigure}
  \hfill
  \begin{subfigure}[b]{0.53\linewidth}
  \centering
        \includegraphics[width=1.0\linewidth]{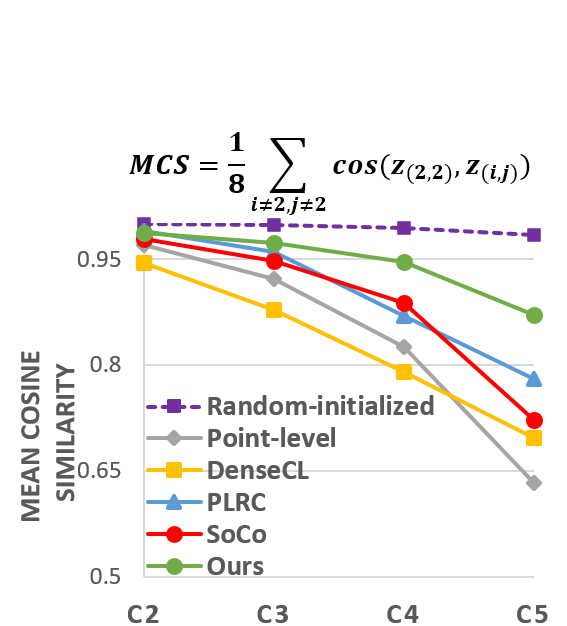}
    \caption{}
    \label{fig:position-b}
  \end{subfigure}
  \vspace{-0.3cm}
  \caption{\textbf{Measuring positional encoding.} (a) We place an image into a $3 \times 3$ grid. (b) We plot the mean cosine similarity between the central grid and the others in different network layers. The similarity decreases with the layer depth, showing that the boundary sub-images are increasingly affected by the positional bias arising from zero-padding.}
 \vspace{-0.5cm}
  \label{fig:pos}
\end{figure}

\vspace{-0.1cm}
The positional bias has been encoded by the zero-padding in CNNs during the training and testing stage, which has been observed in recent works~\cite{xu2021positional,islam2021position}. However, none of them address this issue in the context of dense prediction with multiple objects. To study the influence of the positional bias arising from zero-padding in each layer, we synthesize images where the same object is repeated in a $3 \times 3$ grid, as shown in Fig.~\ref{fig:position-a}. We extract features for each of these 9 objects by forward passing them through networks trained with different SSL methods. Finally, we report the mean cosine similarity (MCS) between the central one and the others. Ideally, this value should be close to 1, as the image patches are identical. However, Fig.~\ref{fig:position-b} shows that deeper layers lead to lower similarity values, indicating that they encode more positional information. The randomly-initialized networks without training have the least positional information, and their results are close to 1 in all the layers. Note that point-level methods encode more positional information than region-level ones. Our D$^3$SSL method, which will be introduced in Section~\ref{sec:method}, yields the lowest coupling rate and positional bias.




\section{Methodology}\label{sec:method}
\begin{figure*}[t]
  \centering
  \includegraphics[width=1.02\linewidth]{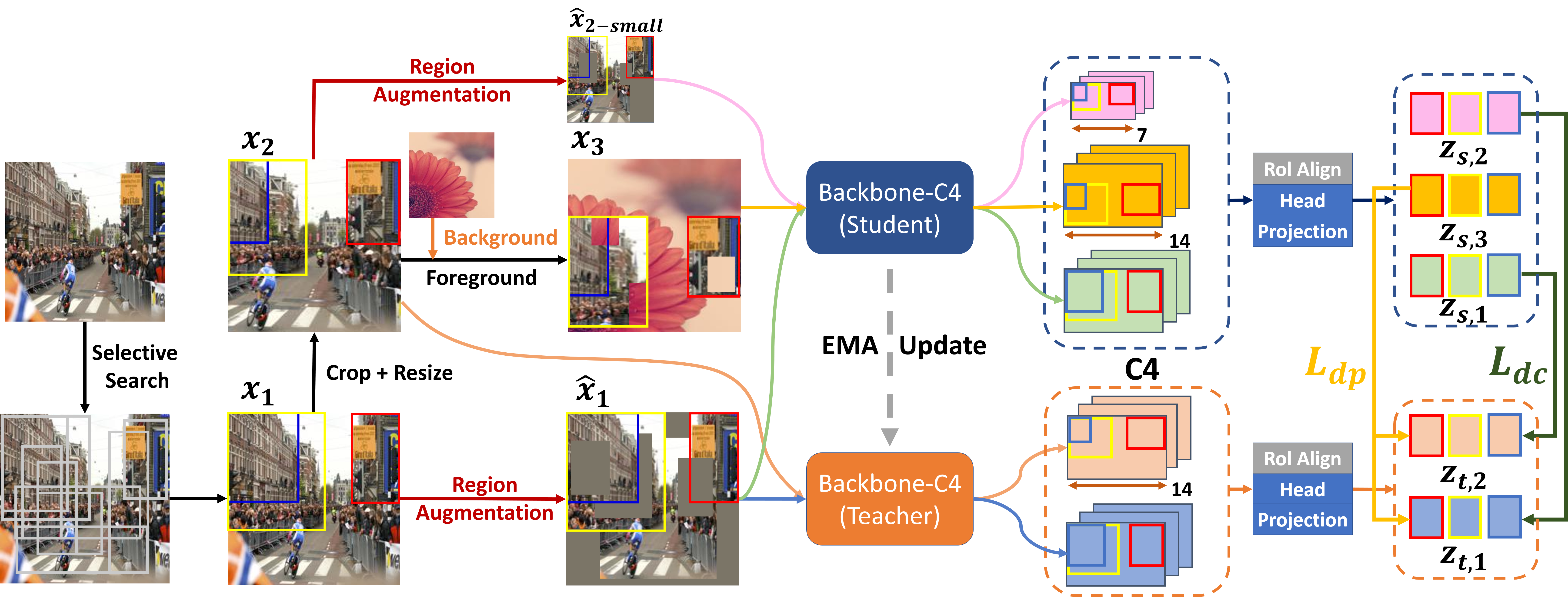}
  \caption{\textbf{Structure of our D$^3$ SSL}. Given an input image, we use external bounding boxes from selective search to perform our three novel region proposal-based augmentations. We then feed the augmented views into the student and teacher backbone and use ROI alignment to obtain the features of the C4 layer. Finally, we calculate our de-coupling and de-positioning losses on the region-level features after the C5 layer.}
  \vspace{-0.6cm}
  \label{fig:structure}
\end{figure*}

The above experiments have unveiled that the current networks have object coupling and positional bias issues. We introduce our approach to addressing them via region-level augmentations and a de-coupling and de-positioning module. Our approach is illustrated in Fig.~\ref{fig:structure}, and we first introduce some preliminary notations and definitions before discussing our contributions.

\textbf{Views construction.} We resize the input image $\mathbf{x}$ to $224\times224$ to get view $\mathbf{x}_1$. View $\mathbf{x}_2$ is obtained by applying a random resized crop with a scale range of $[0.4,1]$. To encourage scale invariance of the object representation, we downsample $\mathbf{x}_2$ to $112\times112$, which yields view $\mathbf{x}_{2-s}$.

\textbf{Region-level feature extraction.} Similar to SoCo~\cite{wei2021aligning},  we use selective search~\cite{uijlings2013selective} to generate noisy bounding boxes. For fair comparisons, we choose the most widely used ResNet50-C4~\cite{he2016deep} structure as the backbone to extract features. For an augmented view $\mathbf{x}_1$ and a bounding box $b$, the corresponding feature representation $\mathbf{z}$ is extracted as
\begin{equation}
    \mathbf{z} = f_\theta(\mathbf{x}_1,b) := f^H_{\theta_H}(\text{RoIAlign}(f^E_{\theta_E}(\mathbf{x}_1),b)),
\end{equation}
where the head function $f^H_{\theta_H}: \mathbb{R}^{h_4 \times w_4 \times c_4} \xrightarrow[]{} \mathbb{R}^d$ denotes the 5-th residual block followed by average pooling and an MLP projector. RoIAlign~\cite{he2017mask} acts on the C4 layer $f^E_{\theta_E}(\mathbf{x})$.



\subsection{Region-level Augmentations}
To reason at the level of regions, following~\cite{henaff2021efficient,wei2021aligning}, we leverage external unsupervised region proposals~\cite{uijlings2013selective} to extract noisy objects. We then introduce three types of novel augmentations and apply them to data as follows. Examples of three augmentations can be found in the supplement.

 \textbf{1) Proposal-based random cutout (PRC):} While randomly resized crops have been heavily used in previous SSL methods, they are ill-suited for region-level SSL due to the different scales and positions of the objects. Hence, we introduce a simple yet effective random cutout strategy based on the proposals to remove partial information.  Our PRC operates from large to small proposals and follows two main design principles. i) Cutout regions are sampled to maintain the same height and width ratio as the proposals, with their sizes randomly chosen from a pre-defined range $[0.2, 0.5]$. ii) The cutout regions of one proposal cannot cover another proposal by more than a threshold. Otherwise, the overlapped cutout region will be recovered to image for smaller proposals. Thus, our PRC guarantees that the cutout regions fit the proposals' shape and prevents removing too much information from the proposals. 
 
 \textbf{2) Proposal-based random mask (PRM):} To extract diverse nearby information with limited box jitter, we propose to randomly add partial background information from different directions of the proposal region. Note that the rectangular masks' scale and aspect ratio are randomly chosen from a range $[0.2,0.6]$ and $[3/4,4/3]$.
 
 \textbf{3) Restricted box jitter (RBJ):} Box jitter applies a random scaling to the width and height of the proposal and shifts its center. However, due to the selective search strategy, larger proposals typically include wider backgrounds and other objects. To clean the proposal regions, the augmented boxes should not be larger than the original ones.

\subsection{Learning Modules}

Learning strategies such as MoCo and BYOL do not take into account the variance of the crops~\cite{zhang2022leverage}, which degrades the learning performance. In region-level training, the object-related information carried by different proposals also varies. Therefore, we treat the proposals differently and put them into the two different training modules depicted in Fig.~\ref{fig:structure}. The de-coupling module aims to capture object features that are resistant to the background and to nearby objects. By contrast, the de-positioning module focuses on addressing the positional bias by compositing the shifted box proposal with an arbitrary background, thus further alleviating coupling. 



\textbf{De-coupling.}
In single-object images, augmentations enforce the background to have the same representation as foreground objects. However, in the presence of multiple objects, we need to prevent the representation of different objects from collapsing into a unified one. Specifically, the proposal-level features can be influenced by visual cues outside of the bounding box since the receptive field always goes beyond the box region. 
To prevent this, we use PRM to incorporate extra background information from different directions into the proposal. In parallel, we also employ PRC to cut a part of the object information. By maximizing the similarity of the resulting two augmented region-level views, the network learns to encode the consensus region of the two views while being resilient to occlusion and the influence of other objects outside the box. 
 Concretely, given a set of $K$ object proposals $\{b_i\}$ in image $\mathbf{x}$, the augmented view $\mathbf{x}_1$ and $\mathbf{x}_{2-s}$ 
can be expressed as
\begin{equation}
\begin{split}
        \hat{\mathbf{x}}_1 &= PRM\circ PRC(\mathbf{x}_1,\{b_i\}), \\
        \hat{\mathbf{x}}_{2-s} &= PRM\circ PRC(\mathbf{x}_{2-s},\{b_i\}).
\end{split}
\end{equation}
Then, the features of each bounding box $b_i$ in both views are extracted from the student and teacher branch as
\begin{equation}
\begin{split}
\mathbf{z}_{s,1}^{(i)}=f_{\theta_q}(\hat{\mathbf{x}}_1,b_i)&,\;\;\mathbf{z}_{s,2}^{(i)}=f_{\theta_q}(\hat{\mathbf{x}}_{2-s},b_i),  \\
\mathbf{z}_{t,1}^{(i)}=f_{\theta_k}(\hat{\mathbf{x}}_1,b_i)&,\;\;\mathbf{z}_{t,2}^{(i)}=f_{\theta_k}(\mathbf{x}_2,b_i),
\end{split}
\end{equation}
where, following the MoCo~\cite{he2020momentum} formalism,  $f_{\theta_q}$ is the online encoder, and  $f_{\theta_k}$ is the momentum encoder. We therefore write our de-coupling loss for bounding box $b_i$ as
\begin{equation}
\begin{split}
\mathcal{L}^{(i)}_{dc}=\mathcal{L}_{nce}(\mathbf{z}_{s,1}^{(i)},\mathbf{z}_{t,2}^{(i)},\{\mathbf{z}^-\})+\mathcal{L}_{nce}(\mathbf{z}_{s,2}^{(i)},\mathbf{z}_{t,1}^{(i)},\{\mathbf{z}^-\}),
\label{eqn:g-to-g}
\end{split}
\end{equation}
where the negative features $\{\mathbf{z}^-\}$ come from a memory bank storing features of augmented regions in previous batches. The contrastive loss $\mathcal{L}_{nce}$ is written as
\begin{small}
\begin{equation}
    \mathcal{L}_{nce}(\mathbf{z},\mathbf{z}^+,\{\mathbf{z}^-\})=-\log\frac{\exp(\mathbf{z}\cdot \mathbf{z}^+/\tau)}{\sum_{\mathbf{z}_j\in \{\mathbf{z}^-\}\cup \{\mathbf{z}^+\}}{\exp(\mathbf{z}\cdot \mathbf{z}_j/\tau)}}.
\end{equation}
\end{small}

\begin{table*}[t]
\centering
\begin{tabular}{c|lll|lll|lll}
    \toprule
                & \multicolumn{6}{c|}{Object Detection}                                                              & \multicolumn{3}{c}{Segmentation}                      \\ \cline{2-10}
                & \multicolumn{3}{c|}{VOC07+12}                    & \multicolumn{3}{c|}{MS-COCO}                        & \multicolumn{3}{c}{MS-COCO}                         \\ \cline{2-10}
                & \multicolumn{1}{c}{$\text{AP}$}          & \multicolumn{1}{c}{$\text{AP}_{50}$}             & \multicolumn{1}{c}{$\text{AP}_{75}$}          & \multicolumn{1}{c}{$\text{AP}$}          & \multicolumn{1}{c}{$\text{AP}_{50}$}             & $\text{AP}_{75}$           & \multicolumn{1}{c}{$\text{AP}$}           & \multicolumn{1}{c}{$\text{AP}_{50}$}             & \multicolumn{1}{c}{$\text{AP}_{75}$}         \\ \hline
MoCo v2\cite{chen2020improved}\scriptsize{\textit{(CVPR20)}} & 54.7\Dn{2.1} & 81.0\Dn{0.7}           & 60.6\Dn{2.9}           & 38.5\Dn{1.1} & 58.1\Dn{1.4}             & 42.1\Dn{1.1}           & 34.7\Up{0.3}          &56.5\Up{0.2}            & 36.9\Dn{0.2}           \\
BYOL\cite{grill2020bootstrap}\scriptsize{\textit{(NeurIPS20)}}            & -         & -          & -          &37.9\Dn{1.7}	&57.5\Dn{2.0}	&40.9\Dn{2.3}
          & -          & -          & -          \\ 
Self-EMD\cite{liu2020self}\scriptsize{(axriv21)}      & -         & - & - &38.5\Dn{1.1}	&58.3\Dn{1.2}	&41.6\Dn{1.6}  & - & - &- \\
PixPro\cite{xie2021propagate}\scriptsize{\textit{(CVPR21)}} &56.5\Dn{0.3}	&81.4\Dn{0.3}	&62.7\Dn{0.8}	&39.0\Dn{0.6}	&58.9\Dn{0.6}	&43.0\Dn{0.2}	&35.4\Up{1.0}	&56.2\Dn{0.1}	&38.1\Up{1.0} \\

 ReSim\cite{xiao2021region}\scriptsize{\textit{(CVPR21)}}	&56.6\Dn{0.2}	&81.7	&63.5\Up{0.5} &39.2\Dn{0.4} &59.1\Dn{0.4}	&42.7\Dn{0.5}	&35.2\Up{0.8}	&56.3	&37.8\Up{0.7}
 \\
DenseCL\cite{wang2021dense}\scriptsize{\textit{(CVPR21)}}	&56.7\Dn{0.1}	&81.7	&63.0\Dn{0.5}	&39.6	&59.3\Dn{0.2}	&43.3\Up{0.1}	&35.7\Up{1.3}	&56.5\Up{0.2}	&38.4\Up{1.3} \\

SoCo\cite{wei2021aligning}\scriptsize{\textit{(NeurIPS21)}}	&\underline{56.8}	&\underline{81.7}	&\underline{63.5} &\underline{39.6}	&\underline{59.5}	&\underline{43.2}	&\underline{34.4}	&\underline{56.3}	&\underline{37.1}
 \\
PLRC\cite{bai2022point}\scriptsize{\textit{(CVPR22)}}	&57.1\Up{0.3}	&82.1\Up{0.4}	&\textbf{63.8}\Up{0.3}	&39.8\Up{0.2}	&59.6\Up{0.1}	&43.7\Up{0.5}	&\textbf{35.9}\Up{1.5}	& \textbf{56.9}\Up{0.6}	&\textbf{38.6}\Up{1.5} \\

\textbf{D$^3$SSL } &\textbf{57.8}\Up{1.0}	&\textbf{82.5}\Up{0.7}	&\textbf{64.4}\Up{0.9}	&\textbf{40.3}\Up{0.7}	&\textbf{60.1}\Up{0.6}	&\textbf{44.0}\Up{0.8}	&35.1\Up{0.7}	&\textbf{56.9}\Up{0.6}	&37.6\Up{0.5} \\
\bottomrule
\end{tabular}
\vspace{-0.2cm}
\caption{\textbf{Main results with COCO pre-training for 800 epochs.} From left to right, we show the average precision on masks and bounding boxes on VOC detection, COCO object detection, and COCO semantic segmentation. The comparison is made with the baseline method, SoCo, which is underscored in the table. We report the percentage of performance decreases and increases for other methods using small red and blue numbers, respectively. The best scores are in \textbf{bold}.} \label{tab:coco 800}
\vspace{-0.4cm}
\end{table*}

\textbf{De-positioning.}
The above operations aim to make the network robust to the object coupling issue but do not handle the positional bias.
To achieve this, we blend the proposal regions in another image and at another location to enforce the robustness of the representation to the positioning and further reduce the entangling with the original background. Specifically, we first add the same random bias to all the bounding boxes $\hat{b}_i = b_i + t$, ensuring that the maximum and minimum coordinates do not exceed the image size. Then, we employ the proposal regions $\{ b_i\}$ as foreground and place them on a randomly sampled background image according to $\{ \hat{b}_i\}$. Specifically, denoting $B(b_i)$ the region covered by bounding box $b_i$, we first define the foreground mask $\mathbf{m}$ as
\begin{equation}
\mathbf{m}(i, j)=\left\{\begin{array}{ll}
1, & \text { if $(i,j) \in B(b_1)\cup ... \cup B(b_K)$}\\
0, & \text { else. } 
\end{array}\right.
\end{equation}
Similarly, we can obtain the translated mask $\hat{\mathbf{m}}$ according to $\hat{b}_i$. As ROI pooling~\cite{he2017mask,ren2015faster} extracts the features in block C4, the C5 block can only access the features within the ROI. Therefore, the foreground masks need an extra processing to incorporate the background information in the C5 block. To this end, we cut the foreground mask again using our PRC as
\begin{equation}
\hat{\mathbf{m}} = PRC(\hat{\mathbf{m}},\{\hat{b}_i\}).
\end{equation}
Finally, we define a composition operation $C$. It copies the foreground region of image $\mathbf{x}_2$ corresponding to the mask $\mathbf{m}$ and pastes it onto a random background image $\mathbf{x}_b$ with a shifted mask $\hat{\mathbf{m}}$, accounting for removing positional bias with a shift value $t$. It returns the composited view, and its features from each bounding box can be extracted as:
\begin{equation}
\mathbf{x}_3 = C(\mathbf{x}_2,\mathbf{m}, \mathbf{x}_b,\hat{\mathbf{m}}), \quad
\mathbf{z}^{(i)}_{s,3} = f_{\theta_q}(\mathbf{x}_3,\hat{b}_i).
\end{equation}


    


The operations above affect the proposal region in terms of both position and surrounding information. Therefore, each feature $\mathbf{z}^{(i)}_{s,3}$ typically encodes different positions and background information from de-coupling ones. Hence, we treat it as the local features of the proposal and forward it to the student branch to push them towards the global features, which allows the representations to be robust to positioning and coupling. We thus write the de-positioning loss as

\vspace{-0.2cm}
\begin{equation}
\mathcal{L}^{(i)}_{dp}=\mathcal{L}_{nce}(\mathbf{z}^{(i)}_{s,3},\mathbf{z}_{t,2}^{(i)},\{\mathbf{z}^-\})+\mathcal{L}_{nce}(\mathbf{z}^{(i)}_{s,3},\mathbf{z}_{t,1}^{(i)},\{\mathbf{z}^-\}).
\label{eqn:l-to-g}
\end{equation}

Altogether, for an input $\mathbf{x}$ with $K$ proposals bounding boxes, our D$^3$SSL loss can be formulated as
\begin{equation}
\mathcal{L}_{D^3SSL}=\frac{1}{K}\sum_{i=1}^K({\mathcal{L}_{dp}^{(i)}+\mathcal{L}_{dc}^{(i)}}).
\end{equation}

\section{Main Empirical Results}\label{sec:experiments}

As we target SSL in the wild, we perform our  SSL training on datasets with multiple objects, namely COCO and our own OpenImage-MINI described below. Following the protocol of~\cite{bai2022point, wang2021dense}, we pre-train our model on COCO and on OpenImage-MINI for 800 epochs. We defer all the training details, such as learning rate, batch size, and augmentation hyper-parameters to the supplementary material.
\begin{table}[h]
\centering
\begin{tabular}{ccccc}
\hline
 & \#Imgs & \#Obj/Img & \#Cat/Img & \#Cat \\ \hline
\begin{tabular}[c]{@{}c@{}}\small OpenImage-MINI\end{tabular} & 146K & 9.1 & 3.20 & 569 \\
COCO & 118K & 7.4 & 2.95 & 80 \\ \hline
\end{tabular}
\vspace{-0.2cm}
\caption{\textbf{Comparison of statistics between two datasets}. From left to right, it shows the number of images (Imgs), the average number of objects per image (Obj/Img) and the average number of categories per image (Cat/Img), and the total number of categories (Cat). }
\label{tab: dataset stat}
\vspace{-0.5cm}
\end{table}

\subsection{OpenImage-MINI Dataset}


To best evaluate dense SSL methods, instead of pretraining models on a simple scene and curated dataset like ImageNet~\cite{russakovsky2015imagenet}, we aim to perform SSL in the wild.
Hence, we create a new dataset called OpenImage-MINI, which is a subset of OpenImage~\cite{OpenImages}. With $\sim$9M annotated images and 600 boxable categories, OpenImage is close to natural class statistics and avoids inductive bias. To ensure a fair comparison of pre-training, we follow the label distribution of COCO~\cite{lin2014microsoft} when sampling the data. Specifically, we select images based on the object annotations in COCO, with the following criteria: 1) at least one object in the image must be from a class in COCO, which will include more semantic categories, and 2) the number of objects from shared categories should not exceed it is in COCO. Using these guidelines, we obtained a subset of approximately 146k images with shared labels that have a similar distribution to COCO. Furthermore, as shown in Table~\ref{tab: dataset stat}, our dataset includes objects seven times more categories than those from COCO, and the average number of objects per image is higher than that in COCO.

\subsection{Dense Prediction}

\textbf{Setting.}
Our evaluation setting on COCO~\cite{lin2014microsoft} and VOC~\cite{everingham2010pascal} follows one of the protocols in MoCo~\cite{he2020momentum}. Specifically, we adopt the ResNet50-C4 backbone. For VOC detection, we fine-tune a Faster R-CNN structure~\cite{ren2015faster} on the combined set of \textit{trainval2007} and
\textit{trainval2012}, and report AP, AP50, and AP75 on the \textit{test2007} set. For COCO detection and segmentation, we adopt a Mask R-CNN~\cite{he2017mask} structure with a $1 \times$ schedule. 

\begin{table*}[h]
\begin{tabular}{c|llllll|lll}
\hline
\multicolumn{1}{l|}{} & \multicolumn{6}{c|}{Object Detection} & \multicolumn{3}{c}{Segmentation} \\ \cline{2-10} 
\multicolumn{1}{l|}{} & \multicolumn{3}{c|}{VOC07+12} & \multicolumn{3}{c|}{MS-COCO} & \multicolumn{3}{c}{MS-COCO} \\ \cline{2-10} 
\multicolumn{1}{l|}{} & \multicolumn{1}{c}{$\text{AP}$} & \multicolumn{1}{c}{$\text{AP}_{50}$} & \multicolumn{1}{c|}{$\text{AP}_{75}$} & \multicolumn{1}{c}{$\text{AP}$} & \multicolumn{1}{c}{$\text{AP}_{50}$} & \multicolumn{1}{c|}{$\text{AP}_{75}$} & \multicolumn{1}{c}{$\text{AP}$} & \multicolumn{1}{c}{$\text{AP}_{50}$} & \multicolumn{1}{c}{$\text{AP}_{75}$} \\ \hline
MoCo v2\cite{chen2020improved}\scriptsize{\textit{(CVPR20)}} & 54.5\Dn{2.4}  & 81.0\Dn{0.9}  & \multicolumn{1}{l|}{60.2\Dn{3.3}} & 38.1\Dn{0.3}  & 57.4\Dn{0.3}  & 41.1\Dn{0.4}  & 33.1\Dn{0.3}  & 54.3\Dn{0.3}  & 35.3\Dn{0.1} \\
DenseCL\cite{wang2021dense}\scriptsize{\textit{(CVPR21)}} & 57.1\Up{0.2} & 82.0\Up{0.1} &  \multicolumn{1}{l|}{63.3\Dn{0.2}} & 39.0\Up{0.6} & 58.6\Up{0.8} & 42.0\Up{0.5} & 34.0\Up{0.6} & 55.2\Up{0.6} & 36.5\Up{1.1} \\
ReSim\cite{xiao2021region}\scriptsize{\textit{(CVPR21)}} & 57.4\Up{0.6} & 82.4\Up{0.6} & \multicolumn{1}{l|}{64.0\Up{0.5}} & 38.9\Up{0.5} & 58.1\Up{0.3} & 42.0\Up{0.6} & 33.9\Up{0.5} & 54.8\Up{0.2} & 36.2\Up{0.8} \\
SoCo\cite{wei2021aligning}\scriptsize{\textit{(NeurIPS21)}} & \underline{56.9} & \underline{81.9} & \multicolumn{1}{l|}{\underline{63.5}} & \underline{38.4} & \underline{57.8} & \underline{41.5} & \underline{33.4} & \underline{54.6} & \underline{35.4} \\
PLRC\cite{bai2022point}\scriptsize{\textit{(CVPR22)}} & 53.7\Dn{3.2} & 80.0\Dn{1.9} & \multicolumn{1}{l|}{58.8\Dn{4.7}} & 36.5\Dn{1.9} & 55.2\Dn{2.6} & 38.8\Dn{2.7} & 31.7\Dn{1.7} & 51.9\Dn{2.7} & 33.7\Dn{1.7} \\
\textbf{D$^3$SSL } & \textbf{57.9}\Up{1.0} & \textbf{82.8}\Up{0.9} & \multicolumn{1}{l|}{\textbf{64.5}\Up{1.0}} & \textbf{40.1}\Up{1.7} & \textbf{59.9}\Up{2.1} & \textbf{43.4}\Up{1.9} & \textbf{34.8}\Up{1.4} & \textbf{56.5}\Up{1.9} & \textbf{36.8}\Up{1.5} \\ \hline
\end{tabular}
\vspace{-0.2cm}
\caption{\textbf{Main results with pre-training on OpenImage-MINI for 800 epochs}. We show the performance of VOC detection, COCO object detection, and semantic segmentation. We compare with the methods that are open-source. The comparison is made with the baseline method, SoCo, which is underscored in the table. We report the percentage of performance decreases and increases for other methods using small red and blue numbers, respectively. The best scores are in \textbf{bold}.} \label{tab:OpenMINI 800}
\vspace{-0.4cm}
\end{table*}

\textbf{COCO pre-training.}
Tab.~\ref{tab:coco 800} compares our D$^3$SSL to the state-of-the-art SSL methods on three downstream dense prediction tasks. Note that all models are pre-trained on COCO for 800 epochs, and we report the results based on the official checkpoints or implementations. We compare three types of methods according to how the representations are extracted in the pre-training stage: 1) Image-level, $i.e.$ the representations are extracted from the whole feature map by average pooling, including MoCo v2~\cite{chen2020improved}, BYOL~\cite{grill2020bootstrap}; 2) Point-level, $i.e.$ the feature vectors without pooling are used for learning, including Self-EMD~\cite{liu2020self}, PixPro~\cite{xie2021propagate}, DenseCL~\cite{wang2021dense} and PLRC~\cite{bai2022point}; 3) Region-level, which uses a set of features from the feature map followed by average pooling, such as SoCo~\cite{wei2021aligning}.

Our D$^3$SSL outperforms the SOTA methods on both VOC and COCO detection tasks, with a significant improvement in detection tasks. More results can be found in the supplementary material, showing that D$^3$SSL achieves even better results than most methods pre-trained on ImageNet. Note that ImageNet is 10 times larger but simpler than COCO. This finding indicates the possibility of pre-training on the downstream datasets to avoid the domain gap issue. Interestingly, our results on segmentation show that point-level methods typically achieve higher performance than region-level. We conjecture that the point-level methods encode more positional information (see Section~\ref{sec:positioning}) and are thus better suited to test data that has a similar distribution to the pre-training one (both pre-training and testing on COCO). However, D$^3$SSL still yields a significant improvement in segmentation over SoCo with the same R50-C4 structure.

\textbf{OpenImage-MINI pretraining.} Let us now compare our D$^3$SSL model to the state-of-the-art SSL methods pre-trained on the OpenImage-MINI dataset, which contains more diverse objects and categories. We pre-train all methods for 800 epochs with their official codes and report the performance on the VOC and COCO downstream tasks. Note that all the methods use the same hyperparameters as their pre-training on COCO. We pick DenseCL and PLRC as they are the best two among the point-level methods pre-trained on COCO. We also report the results of MoCo~\cite{he2020momentum} as it performs better than other image-level methods in dense prediction tasks. Tab.~\ref{tab:OpenMINI 800} provides the mask and bounding box average precisions of the three tasks. 

Our D$^3$SSL shows strong performance on all the benchmarks, and even achieves better results than all point-level methods on the segmentation task in the presence of a domain gap (from OpenImage to COCO). Comparing the results in Tab.~\ref{tab:coco 800} and Tab.~\ref{tab:OpenMINI 800}, we can observe that the performance of the other methods degrades significantly, e.g.,\ SoCo's AP drops from 39.6 to 38.4 on detection, whereas D$^3$SSL only yields $0.2\%$ decrease. This confirms the results provided in Sec.~\ref{sec:ana}, where our D$^3$SSL achieves the lowest positional bias and coupling rate, and implies that D$^3$SSL has \textbf{better generalization} ability thanks to our de-positioning and de-coupling modules. 

\begin{table}[h]
\centering
\begin{tabular}{c|cc|cc}
    \toprule
                & \multicolumn{2}{c|}{O-KNN(\%)}          & \multicolumn{2}{c}{Disturbed O-KNN(\%)}                          \\ 
                & Top-1  & Top-5 
                & Top-1 & Top-5 \\ \hline
MoCo v2\cite{chen2020improved} 	&68.3 	&86.5	&53.1 	&72.0  \\
DenseCL\cite{wang2021dense}		&65.2 	&86.1 	&44.8 &64.5 	 \\
ReSim\cite{xiao2021region}		&63.3 	&84.6 	&45.4 	&65.8  \\
SoCo\cite{wei2021aligning}		&70.2 	&88.3 	&47.0 	&66.1 	\\
\textbf{D$^3$SSL}		&\textbf{73.1} 	&\textbf{90.1} 	&\textbf{55.8} 	&\textbf{73.4} 	 \\
\bottomrule
\end{tabular}
\vspace{-0.2cm}
\caption{Object-KNN results of models pre-trained for 800 epochs on OpenImage-MINI. We show the Top-1 and Top-5 accuracy of O-KNN and disturbed O-KNN.}
\label{tab:DenseKNN}
\vspace{-0.5cm}
\end{table}

\subsection{Dense-level KNN}

\textbf{Object-level KNN.}
In addition to evaluating our results by fine-tuning for the downstream tasks with supervision, we propose a new dense-level feature evaluation method inspired by K-Nearest Neighbor, namely Object-level KNN (O-KNN). Concretely, we extract object-level features from the C5 block with ground-truth bounding boxes by using RoI Align with a size of $7 \times 7$ and average pooling. For the training set, we extract $N$ object features per image to store the features and labels in a memory bank. Choosing all the objects in the dataset will lead to imbalanced numbers across the categories in the training set, which will lead the results to be biased. Similarly to image-level KNN, we predict the label for each object in the evaluation set by finding the $k$-nearest object-level features in the training set. Thus, the process does not need further training and is much faster. More importantly, compared to the evaluation with a fine-tuning stage, O-KNN  directly reflects how well the features from the same class are preserved instead of favoring the mask or detection prediction. Additional visualizations and analyses on the coupling of features among objects and backgrounds can be found in the supplementary materials.


\textbf{Disturbed Object-level KNN.}
To quantify the influence of coupling for multi-object images, based on our O-KNN, we further conduct O-KNN with background noise introduced. Specifically, the feature extraction pipeline is kept the same, while all background regions, $i.e.$ regions outside the selected ground-truth bounding boxes, are replaced by a randomly sampled image. This setting helps us to further explore how irrelevant background information can affect the dense-level recognition ability of a pre-trained network.

\textbf{O-KNN Comparison on COCO.} Tab.~\ref{tab:DenseKNN} shows the  O-KNN and disturbed O-KNN accuracy on COCO with models pre-trained on OpenImage-MINI for 800 epochs, and using COCO \textit{train2017} for object feature extraction and \textit{val2017} for evaluation. D$^3$SSL outperforms the other methods by a large margin on both the O-KNN and disturbed O-KNN tasks. Compared to the other dense-level methods, MoCo is more robust to the background information. This again indicates that the current dense-level SSL methods highly rely on the correlation of objects and position bias. By contrast, thanks to our augmentation techniques and two modules, D$^3$SSL alleviates such disturbances.


\section{Ablation Study}\label{sec:ablation}

In this section, we dissect our model and study the impact of each part. 
 \vspace{-0.3cm}
\begin{table}[h]
\centering
\begin{tabular}{cc|ccc|l}
\toprule
\begin{tabular}[c]{@{}c@{}}Random\\ Cutout\end{tabular} & \begin{tabular}[c|]{@{}c@{}}Box \\ Jittering\end{tabular} &PRC  & RBJ & PRM & \multicolumn{1}{c}{AP} \\\cline{1-6}
 &  &  &  &  & 54.5 \\
 $\surd$&  &  &  &  & 54.6\Up{0.1} \\
 &$\surd$  &$\surd$  &  &  & 54.9\Up{0.4} \\
 &  &$\surd$  &  &  & 54.9\Up{0.4} \\
 &  &$\surd$  &$\surd$  &  & 55.2\Up{0.7}  \\
 &  &$\surd$  &$\surd$  &$\surd$  & 55.5\Up{1.0} \\
\bottomrule
 \end{tabular}
 \vspace{-0.2cm}
 \caption{\textbf{Ablation studies of the combination of augmentation methods}. The leftmost two augmentations are used in SoCo, the middle three are our methods, and the AP denotes the average precision (the higher, the better) on VOC detection. A \checkmark denotes that the augmentation is used on top of our method with the decoupling branch.} \label{tab:ablation_aug}
 \vspace{-0.3cm}
\end{table}

\textbf{Our Augmentations.} To evaluate the effect of our region-level augmentations, we pretrain on COCO \textit{train2017} for 200 epochs without $\mathcal{L}_{dp}$ due to simplicity. Tab.~\ref{tab:ablation_aug} reports the AP on VOC detection of different combinations of region-level augmentation. The baseline applies no region-level augmentation. By comparing the results on all rows, we observe that the PRC operation yields better performance (+0.4\%) than the original cutout in the region-level SSL. This shows that our proposed random cut is more suitable for dense-level SSL. However, box jittering does not contribute to our PRC, indicating that inconsistency of augmentations on the bounding box does not help the optimization. By contrast, our RBJ and PRM further boost the performance without causing any conflicts. Compared to the baseline, applying our three augmentations yields +1.0 AP improvement. Moreover, we apply our augmentation methods to the SoCo framework, and the results in Tab.~\ref{tab:ablation SoCo} demonstrate that our augmentation methods generalize to other region-level frameworks.

\begin{table}[h]
\centering
\resizebox{\linewidth}{!}{
\begin{tabular}{c|lll|l}
 & AP & AP$_{50}$ & AP$_{75}$ & CR $\downarrow$ \\ \hline
SoCo & 55.3 & 80.4 & 61.6 & 1.31 \\
SoCo+Aug$_{D^3SSL}$ & 55.8\Up{0.5} & 81.2\Up{0.8} & 62.1\Up{0.5} & 1.20 \\ \hline
\end{tabular}
}
\vspace{-0.2cm}
\caption{\textbf{Applying our augmentations to SoCo}. The two models are trained for 200 epochs on OpenImage-MINI, and fine-tuned and tested on VOC (07+12).} 
\vspace{-0.4 cm}
\label{tab:ablation SoCo}
\end{table}

\textbf{Modules.}
To study how both the modules contribute to our model, we start with training only with the $\mathcal{L}_{dc}$ loss, denoted as w/o $\mathcal{L}_{dp}$.
We also pre-train D$^3$SSL with the full setting on COCO for 200 epochs and evaluate on VOC. As shown in Tab.~\ref{tab:dd}, our full model with $\mathcal{L}_{dp}$ yields a $+1.5\%$ AP improvement compared to the model without $\mathcal{L}_{dp}$. Furthermore, we compute the coupling rate and mean cosine similarity for both backbones to test the coupling and positioning issues, following the setting in $\xi$\ref{sec:coupling} and $\xi$\ref{sec:positioning}. The results show that our de-positioning module helps in both metrics, especially for the positioning problem. Altogether, our framework helps the training to reduce the bias from the nearby objects and from the position.

\begin{table}[h]
\centering
\begin{tabular}{c|cc|l}
\hline
 & \begin{tabular}[c]{@{}c@{}}CR $\downarrow$ \end{tabular} & \begin{tabular}[c]{@{}c@{}} MCS $\uparrow$ \end{tabular} & \multicolumn{1}{c}{AP $\uparrow$}  \\ \hline
w/o $\mathcal{L}_{dp}$ & 1.33 & 0.68 & 55.5 \\
w/ $\mathcal{L}_{dp}$ & 1.17 & 0.82 & \textbf{57.0}\Up{1.5} \\ \hline
\end{tabular}
\vspace{-0.2cm}
 \caption{\textbf{Ablation study of our training modules}. We follow the procedure in Sec.~\ref{sec:ana} to construct images from OpenImage-MINI, while our models w/ $\mathcal{L}_{dp}$ and w/o $\mathcal{L}_{dp}$ are pre-trained on COCO for 200 epochs. We also report the AP of both models on VOC detection tasks. } \label{tab:dd}
\vspace{-0.6cm}
\end{table}

\section{Conclusion and Limitations}\label{sec:con}
In this paper, we have presented a novel dense SSL framework, namely D$^3$SSL, to tackle the problem of self-supervised learning in complex scenes and dense-level feature learning. To that end, we have introduced region-level augmentation techniques and corresponding two modules, de-coupling and de-positioning, to enable the network to learn features that are robust to position and background changes, which has been verified by our experiments. Moreover, we have introduced two metrics and experiments to measure the positioning and coupling quantitatively and created a new dataset, OpenImage-MINI. Our extensive experiments have demonstrated the advantages of D$^3$SSL on several standard benchmarks, taking a step further toward deploying SSL in the wild. We hope that this study will inspire and attract the community's attention to object-level SSL in real-world datasets.

\textbf{Limitations.} Our D$^3$SSL needs external information such as proposals from selective search to start with, which incurs additional computation compared to image-level SSL. Besides, it may favor the detection task over the segmentation one. Therefore, we will explore conducting dense-level SSL without extra information.

\small{\textbf{Acknowledgement.} This work was supported in part by the National Natural Science Foundation of China under Grant No. 62006182 and the Swiss National Science Foundation via the Sinergia grant CRSII5-180359.}
{\small
\bibliographystyle{ieee_fullname}
\bibliography{egbib}
}

\end{document}